%% file: sl-single.tex
\documentclass[letterpaper]{article} 

\usepackage{times}  
\usepackage{helvet}  
\usepackage{courier}  
\usepackage{url}  
\usepackage{graphicx}  
\usepackage[utf8]{inputenc} 
\usepackage[T1]{fontenc}    
\usepackage{booktabs}       
\usepackage{amsfonts}       
\usepackage{nicefrac}       
\usepackage{microtype}      
\usepackage{tikz}
\usetikzlibrary{shapes.misc, positioning}
\usepackage{thmtools,thm-restate}
\usepackage{parskip}
\usepackage{authblk}
\usepackage{amsmath}  
\usepackage{amsthm}
\usepackage[numbers,sort]{natbib}

\usepackage{floatrow}
\newfloatcommand{capbtabbox}{table}[][\FBwidth]

\usepackage{hyperref}
\usepackage{cleveref}

\usepackage{macros}
\usepackage{basic}
\usepackage{algorithm}
\usepackage{algorithmic}
\usepackage{tikz}

\makeatletter
\def\thm@space@setup{%
  \thm@preskip=\parskip \thm@postskip=0pt
}
\makeatother

\title{Learning Dependency Structures for Weak Supervision Models}

 \author[$\ddagger$]{Paroma~Varma$^*$}
 \author[$\dagger$]{Frederic~Sala\thanks{Equal Contribution}}
 \author[$\dagger$]{Ann~He}
 \author[$\dagger$]{Alexander~Ratner}
 \author[$\dagger$]{Christopher~R{\'e}}

 \affil[$\dagger$]{Department of Computer Science, Stanford University}
 \affil[$\ddagger$]{Department of Electrical Engineering, Stanford University}
 \affil[ ]{\footnotesize{\texttt{\{paroma, fredsala, annhe, ajratner\}@stanford.edu, chrismre@cs.stanford.edu}}}

\begin{document}

\maketitle

\begin{abstract}
    \input{abstract}
\end{abstract}

  \section{Introduction}
  \label{sec:intro}

\input{intro}

  \section{Background}
  \label{sec:background}

\input{bg}

  \section{Learning Structures in the Weak Supervision Regime}
  \label{sec:theory}

\input{theory}

  \section{Analysis}
  \label{sec:analysis}

\input{analysis}

  \section{Experimental Results}
  \label{sec:exp}

\input{exp}
  \section{Conclusion}
  \label{sec:conc}
  \input{conc}

 \bibliography{sl}
\bibliographystyle{plainnat}
 \appendix
  
\allowdisplaybreaks

  \section{Glossary \& Extended Related Work}
  First, we provide a glossary of terms and notation that we use throughout this paper for easy summary. Afterwards, we provide an extended discussion of related work. We give the proofs of our main results (the lemma and the two theorems). Next, we include a discussion on other aspects of generating information-theoretic lower bounds for the weak supervision setting. We also consider extending robust PCA-based techniques for structure learning \emph{without} the singleton separator set assumption. Finally, we give additional experimental details.

  \label{sec:gloss}
  \input{glossary}

  
  \paragraph{Extended Related Work}
  \input{related}
  
  \section{Proofs}
  \label{sec:app_proofs}
  \input{proofs}

  \section{Extended Experimental Details}
  \label{sec:extexp}
  \input{ext.tex}

\end{document}

%% file: abstract.tex
Labeling training data is a key bottleneck in the modern machine learning pipeline. 
Recent \textit{weak supervision} approaches combine labels from multiple noisy sources by estimating their accuracies without access to ground truth labels; however, estimating the dependencies among these sources is a critical challenge. 
We focus on a robust PCA-based algorithm for learning these dependency structures, establish improved theoretical recovery rates, and outperform existing methods on various real-world tasks.
Under certain conditions, we show that the amount of unlabeled data needed can scale sublinearly or even logarithmically with the number of sources $m$, improving over previous efforts that ignore the sparsity pattern in the dependency structure and scale linearly in $m$.
We provide an information-theoretic lower bound on the minimum sample complexity of the weak supervision setting.
Our method outperforms weak supervision approaches that assume conditionally-independent sources by up to $4.64$ F1 points and previous structure learning approaches by up to $4.41$ F1 points on real-world relation extraction and image classification tasks.

%% file: intro.tex
Supervised machine learning models have increasingly become dependent on a large amount of labeled training data.
For most real-world applications, however, hand labeling such a large magnitude of data is a major bottleneck, especially when domain expertise is required. 
Recently, generative models have been used to combine noisy labels from \textit{weak supervision} sources, such as user-defined heuristics or knowledge bases, to efficiently assign training labels by treating the true label as a latent variable~\cite{alfonseca2012pattern,takamatsu2012reducing,roth2013combining,ratner2016data,Ratner19}.
Once the labels from the multiple noisy sources are used to learn the parameters of a generative model, the distribution over the true labels is inferred and used to produce probabilistic training labels for the unlabeled data, which can then be used to train a downstream discriminative model.

 \begin{figure}[t]
   \centering
  \includegraphics[width=0.8\linewidth]{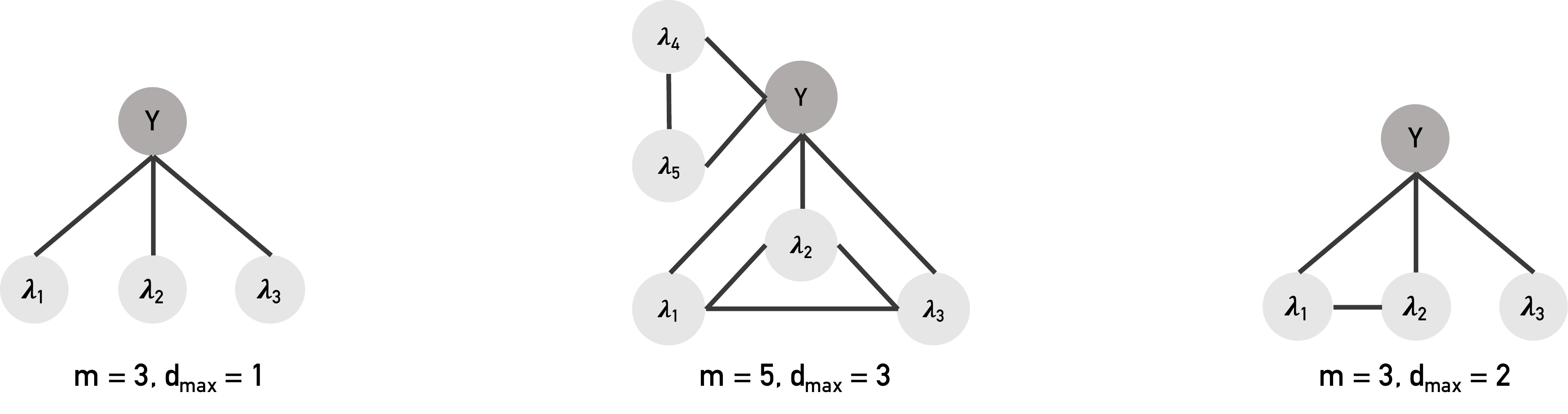}
  \caption{Example graphical structures that can occur in the weak supervision setting and that our method aims to learn. Here, the $\lambda_i$'s represent weak supervision sources and $Y$ represents the latent true variable. Edges indicate dependencies; note that there is always an edge between $\lambda_i$ and $Y$. The number of sources is $m$ while $d_{max}$ is the maximum degree of a source.}
   \label{fig:graphs}
 \end{figure}

Specifying how these weak supervision sources are correlated is essential to correctly estimating their accuracies.
In practice, weak supervision sources often have strongly correlated outputs due to shared data sources or labeling strategies; for example, developers might contribute near-duplicate weak supervision sources.
Manually enumerating these dependencies is a prohibitive development bottleneck, while learning them statistically usually requires ground truth labels~\cite{meinshausen2006high,zhao2006model,ravikumar2010high,Loh13}.
Recently, \citet{bach2017learning} proposed a structure learning method in the weak supervision setting that requires $\Omega(m \log m)$ samples given $m$ sources and \emph{does not exploit the sparsity of the associated model}.
This high sample complexity may prevent it from identifying dependencies, thus affecting the downstream quality of training labels assigned by the generative model.

We propose using a structure learning technique for the weak supervision setting that exploits the sparsity of the model to achieve improved theoretical recovery rates.
We decompose the inverse covariance matrix of the observable sources via robust principal component analysis~\cite{Candes11,Chandrasekaran11}. 
The decomposition produces a sparse component encoding the underlying structure and a low-rank component due to marginalizing over the latent true label variable. We build on previous approaches using this technique~\cite{Chandrasekaran11, Wu17}, but improve over their requirement of $\Omega(m)$ samples under common weak supervision conditions.

The key to obtaining tighter complexity estimates is characterizing the \emph{effective rank}~\cite{Vershynin12} of the covariance matrix in terms of the structural information associated with the weak supervision setting. The effective rank can be unboundedly smaller than the true rank. 
We show that under certain reasonable conditions on the effective rank of the covariance matrix, intuitively similar to the presence of a stronger dependency in each cluster of correlated sources, the sample complexity can be sublinear $\Omega(d^2 m^{\tau})$ for $0 < \tau < 1$ and maximum dependency degree $d$. Under a stronger condition equivalent to the presence of a dominant cluster of correlated supervision sources, we obtain the rate $\Omega(d^2 \log m)$ that matches the \emph{optimal supervised rate}~\cite{Santhanam12}. We further study the unsupervised setting through an information-theoretic lower bound on the sample complexity, yielding a characterization of the \emph{additional cost of the weak supervision setting} compared to the supervised setting. We find that, although latent-variable structure learning may result in much higher sample complexity in general, in the weak supervision setting, the additional number of samples required is small.

For a variety of real-world tasks from relation extraction to image classification, correlations often naturally arise among weak supervision sources like distant supervision via dictionaries and user-defined heuristics. We show that modeling dependencies recovered by our approach improves over assuming conditional independence among the weak supervision sources by up to $4.64$ F1 points, and over existing structure learning approaches by up to $4.41$ F1 points.

%% file: bg.tex
\paragraph*{Related Work}
Manually labeling training data can be expensive and time-consuming, especially in cases requiring domain expertise. 
A common alternative to labeling data by hand is using weak supervision sources. Estimating the accuracies of these sources without ground truth labels is a classic problem~\cite{dawid1979maximum}. Methods like crowdsourcing~\cite{dalvi2013aggregating,joglekar2015comprehensive,zhang2016spectral}, and boosting~\cite{schapire2012boosting} are common approaches; however, we focus on the case in which \emph{no labeled data} is required. Recently, generative models have been used to combine various sources of weak supervision in such settings~\cite{alfonseca2012pattern,takamatsu2012reducing,roth2013combining,ratner2016data,Ratner19}.

Dependencies occur naturally among weak supervision sources for a variety of reasons: sources may operate over the same input~\cite{varma2017inferring}, distant supervision sources may refer to similar information from a single knowledge base~\cite{mintz2009distant}, and heuristics over ontologies may operate over the exact same subtree~\cite{mallory2015large}. While some of these dependencies can be explicit, dependencies are difficult to specify manually in cases with hundreds of sources, potentially developed by many users. Therefore, there is a need to learn dependencies directly from the labels assigned by the weak supervision sources without using ground truth labels.

Structure learning has a rich history outside of the weak supervision setting. The supervised, fully observed setting includes node-wise and matrix-wise methods. Node-wise methods, like~\citet{ravikumar2010high}, use regression on a particular node to recover that node's neighborhood. Matrix-wise methods use the inverse covariance matrix to determine the structure~\cite{Friedman08, Ravikumar11, Loh13}. 
In the latent variable setting, works like \citet{Chandrasekaran12,Meng14, Wu17} perform structure learning via robust-PCA like approaches. In contrast to these works, we focus on the weak supervision setting, providing a tighter characterization that leads to improved rates.
We include further details on related work in the Appendix.

The major work for structure learning in the weak supervision regime is \citet{bach2017learning} which uses a $\ell_1$-regularized node-wise pseudo-likelihood method to obtain a sample complexity of $\Omega(m \log m)$. Note that this expression does not depend on the maximum dependency degree $d$. Our approach fundamentally differs---we use a matrix-wise method that scales better with key parameters (like the sparsity of the graph $d$) and offers improved performance for several real-world tasks. 

\paragraph{Problem Setup}
We formally describe our setup and the generative model we use to assign probabilistic training labels given a set of noisy labels from weak supervision sources. $X \in \mathcal{X}$ is a data point, $\y \in \mathcal{Y}$ is a label with $(X, \y)$ drawn i.i.d. from some distribution $\mathcal{D}$. In the weak supervision setting, we never have access to the true label $\y$; instead we rely on $m$ weak supervision sources that produce noisy labels $\lf_i$ for $1 \leq i \leq m$. 

\begin{example}In a text relation extraction setting, $X$ could be be a tuple of two words, such as names of people, and $\y \in \{0,1\}$ then represents whether the relation of interest exists between the two words, for example whether these two people are being described as married. Potential weak supervision sources can use information from the sentence, such as whether the word ``married'' appears between the two words, to heuristically---and thus noisily---assign a label for a data point $X$. An example of an erring label is produced by applying the heuristic to the sentence “Bob and Alice were meant to get married in 2018, but postponed the wedding by 3 years.”
\end{example}
 
We model the joint distribution of $\lf_1, \lf_2, \ldots, \lf_m, \y$ via a Markov random field with associated graph $G = (V,E)$ with $V = \{\lf_1, \ldots, \lf_m\} \cup \{\y\}$.
If $\lf_i$ is not independent of $\lf_j$ conditioned on $Y$ and the other sources, then $(\lf_i, \lf_j)$ is an edge in $G$. Examples of such graphs are shown in Figure~\ref{fig:graphs}.

For simplicity, we assume $\mathcal{X}, \mathcal{Y} = \{0,1\}$, although our results easily extend. 
The density $f_G$ is then given by
\begin{align}
f_G(\lambda_1,  \ldots,  \lambda_m, y) = \frac{1}{Z}\exp  \left( \sum_{\lambda_i \in V} \theta_{i} \lambda_i \right. \nonumber + \left. \sum_{(\lambda_i,\lambda_j) \in E} \theta_{i,j} \lambda_i \lambda_j  + \theta_Y y + \sum_{\lambda_i \in V} \theta_{Y,i} y \lambda_i \right),
\end{align}


where $Z$ is a partition function to ensure $f_G$ is a normalized distribution, and $\theta_{i}$ and $\theta_{i,j}$ represent the canonical parameters associated with the sources. We can think of $\theta_{i,j}$ as the strength of the correlation between sources $\lf_i$ and $\lf_j$, and $\theta_{Y,i}$ as a measure of accuracy of the source $\lf_i$. Once these parameters are learned, the generative model assigns probabilistic training labels by computing $f_G(Y | \lambda_1, \ldots, \lambda_m)$ for each object $X$ in the unlabeled training set. These probabilistic training labels can then be used to train any downstream discriminative model.

In the conditionally independent model, $\theta_{i,j} = 0$ $\forall i,j$. In cases with dependencies between sources, the structure of $G$ is user-defined or inferred from metadata related to the weak supervision source. 
Once our approach learns the dependency structure, we apply previous work that samples from the posterior of a graphical model directly~\cite{Ratner16,bach2017learning} or uses a matrix completion approach to solve for the accuracy and dependency parameters~\cite{Ratner19}. 

We also rely on a common assumption in weak supervision, the \emph{singleton separator set} assumption~\cite{Ratner19}. This assumption means that our sources form a total of $s$ connected clusters, and is motivated by the intuition that groups of weak supervision sources may share common data resources, core heuristics, or primitives.

%% file: theory.tex
Our goal is to learn the dependency structure among weak supervision sources, i.e. graph $G$, directly from data, without observing the latent true label $\y$. We introduce this \emph{latent structure learning} problem, which we focus on for the remainder of the paper, in Section~\ref{subsec:objective}.
We then provide background on robust PCA in Section~\ref{subsec:rpca}, and describe our algorithm which adapts it to the weak supervision setting in Section~\ref{subsec:implementation}.

\subsection{Structure Learning Objective}
\label{subsec:objective}
We want to learn the structure of graph $G$ given access to noisy labels from $m$ weak supervision sources and no ground truth labels. 
We leverage a common weak supervision assumption that the graph is sparse. This implies that the inverse covariance matrix $\Sigma^{-1}$ of $\lf_1, \ldots, \lf_m, \y$ is \emph{graph-structured}: there is no edge between $\lf_i$ and $\lf_j$ in $G$ when the corresponding term in $\Sigma^{-1}$ is 0, or, equivalently, $\lf_i$ and $\lf_j$ are independent conditioned on all of the other terms~\cite{Loh13}.
However, a key difficulty is that we never know $\y$, so \emph{we cannot observe the full covariance matrix $\Sigma$.}

Let $O = \{\lf_1, \ldots, \lf_m\}$ be the observed labels from the weak supervision sources, and $S = \{\y\}$ be the unobserved latent variable. Then,
\begin{align*}
	\Cov{}{O \cup \mathcal{S} }
	:=
	&~\Sigma
	=
	\begin{bmatrix}
		\Sigma_O & \Sigma_{O\mathcal{S}} \\
		\Sigma_{O\mathcal{S}}^T & \Sigma_\mathcal{S}
	\end{bmatrix}.
\end{align*}

While we cannot observe $\Sigma$ since it contains the true label $\y$, we can observe $\Sigma_O$. Concretely, we form the empirical covariance matrix of observed labels ${\Sigma}_O^{(n)} \in \mathbb{R}^{m \times m}$ in the following manner: $$\Sigma_O^{(n)} = \frac{1}{n}\Lambda \Lambda^T - vv^T$$
where $\Lambda$ represents the $m \times n$ matrix of labels from the weak supervision sources assigned to the unlabeled data, $n$ represents the total number of datapoints, and $v \in \mathbb{R}^{m \times 1}$ is the average label assigned by each of the weak supervision sources. 

We rely on the fact that the inverse covariance matrix
\begin{align*}
	K
	&:=
	\Sigma^{-1}
	=
	\begin{bmatrix}
		K_O & K_{O\mathcal{S}} \\
		K_{O\mathcal{S}}^T & K_\mathcal{S}
	\end{bmatrix}.
\end{align*}
is graph structured~\cite{Loh13}, and therefore so is the sub-block $K_O$. In turn, this implies that $K_O^{-1}$ is a permutation of a block-diagonal matrix with $s$ blocks corresponding to the $s$ source clusters, where each block is no larger than $(d+1) \times (d+1)$, where $d$ is the maximum dependency degree. We will use this fact later on.
From the block matrix inversion formula,
\begin{align}
	\label{eqn:block-inv-cov-main}
	K_O
	&=
	\Sigma_O^{-1}
	+ c\Sigma_O^{-1}\Sigma_{O\mathcal{S}} \Sigma_{O\mathcal{S}}^T\Sigma_O^{-1},
\end{align}
where $c = \left( \Sigma_\mathcal{S} - \Sigma_{O\mathcal{S}}^T\Sigma_O^{-1}\Sigma_{O\mathcal{S}} \right)^{-1} \in \mathbb{R}^+$. Let $z = \sqrt{c} \Sigma_O^{-1}\Sigma_{O\mathcal{S}}$; we can write~(\ref{eqn:block-inv-cov-main}) as 
$$\Sigma_O^{-1} = K_O - zz^T.$$

The empirically observable term $\Sigma_O^{-1}$ is the sum of a graph-structured term ($K_O$) and a rank-one matrix ($zz^T$) that represents marginalizing over the latent label $\y$. Note that while $K_O$ is sparse, adding the dense low-rank matrix $zz^T$ to it will result in $\Sigma_O^{-1}$ being dense as well.


In the latent structure learning setting, our goal is to calculate $K_O$, which is graph-structured and allows us to read off the structure of $G$ from its entries. We therefore have to decompose the observable $\Sigma_O^{-1}$ into $K_O$ and $zz^T$, unknown sparse and low-rank components. This inspires the use of robust principal component analysis~\cite{Candes11,Chandrasekaran11}. 

\subsection{Robust PCA} 
\label{subsec:rpca}
The robust PCA setup consists of a matrix $M \in \mathbb{R}^{m \times m}$ that is equal to the sum of a low-rank matrix and a sparse matrix, $M = L + S$, where $\text{rank}(L) = r$ and $|\supp(S)| = k$. The name of the problem was inspired by the observation that although standard PCA recovers a low-dimensional subspace in the presence of bounded noise, it is not robust to gross corruptions (modeled by the entries of the sparse matrix).
Note that the decomposition $M = L +S$ is not identifiable without additional conditions. For example, if $M = e_ie_j^T$, $M$ is itself both sparse and low-rank, and thus the pairs $(L,S) = (M,0)$ and $(L,S) = (0,M)$ are equally valid solutions. Therefore, the fundamental question of robust PCA is to determine conditions under which the decomposition can be recovered.

The two seminal works on robust PCA \cite{Candes11,Chandrasekaran11} studied \emph{transversality} conditions for identifiability. In particular, the solution spaces  $L, S$ can only intersect at $0$.

For the sparse component, let \[\Omega(S) = \{X \in \mathbb{R}^{m \times m} \text{ }  |  \text{ }  \supp(N) \subseteq \supp(S)\}.\]

For the low-rank component, let $L = UDV^T$ be the SVD of $L$ with rank $r$. Then, let
\begin{align*}
T(L) = \{UX^T + YV^T \text{ } | \text{ } X, Y \in \mathbb{R}^{m \times r} \}.
\end{align*}

The key notion for identifiability in robust PCA problems is to ensure these subspaces are transverse---so that neither the low-rank components are too sparse, nor the sparse component too low-rank. We measure these notions via the the functions $\mu, \xi$ \cite{Chandrasekaran11}:
\[ \mu(\Omega(S)) = \max_{N \in \Omega(S), \|N\|_{\infty} = 1} \|N\|,\]
and
\[ \xi(T(L)) = \max_{N \in T(L), \|N\| \leq 1} \|N\|_{\infty} .\]

These two quantities govern how well-aligned the sparse matrix $S$ is with the coordinate axes and how spread out the low-rank matrix $L$ is. For the decomposition of $M = L+S$ to be identifiable, the required condition is
\begin{align}
\mu(\Omega(S)) \xi(T(L)) < 1.
\label{eq:ident}
\end{align}

\subsection{Adapting Robust PCA for Weak Supervision}
\label{subsec:implementation}
We now adapt the robust PCA setting to our setup: $S = K_O$ and $L = zz^T$, a rank one matrix. First, we determine identifiability in the noiseless case: if we do not have identifiability even with the true $\Sigma_O$ matrix, we have no hope of recovering structure in the sampled case $\hat{\Sigma}_O$. 

Let $a_{\min}, a_{\max}$ be the smallest and largest terms in $\Sigma_{OS}$, respectively. These represent the smallest and largest covariances between the true label $\y$ and the weak supervision sources $\lf_i$, which are the smallest and largest accuracies of the sources. Similarly, we let $c_{\min}, c_{\max}$ be the smallest and largest terms in $\Sigma_O$, respectively, representing the smallest and largest correlations among the sources. We can now write the identifiability condition in terms of the extreme values of the source accuracies and correlations.

\begin{lemma}
Let $K_O$ be the block of the inverse covariance matrix $\Sigma^{-1}$ corresponding to the observed variables, and let $a_{\min}, a_{\max}, c_{\min}, c_{\max}$ be defined as above. Then, 
\[\mu(\Omega(K_O)) \xi(T(zz^T)) \leq \frac{6.4d}{\sqrt{m}} \left(\frac{c_{\max}}{c_{\min}}\right) \left(\frac{a_{\max}}{a_{\min}}\right).\]
\label{lem:ab}
\end{lemma}

Thus, for a fixed degree $d$, if we have access to $$m \geq 40.96d^2 [c_{\max}a_{\max} / (c_{\min}a_{\min})]^2$$ 
weak supervision sources, then $\mu(\Omega(K_O)) \xi(T(zz^T)) < 1$ and there is a unique solution to the decomposition of $\Sigma_O^{-1}$. 

 \begin{algorithm}[tb]
 	\caption{Weak Supervision Structure Learning}
    	\label{alg:sl1}
 	\begin{algorithmic}
 		\STATE \textbf{Input:}
 			Estimate of the covariance matrix $\hat{\Sigma}_O$, parameters $\lambda_n, \gamma$, threshold $T$, loss function $\mathcal{L}(\cdot, \cdot)$

 		\STATE \textbf{Solve:} \[(\hat{S}, \hat{L}) = \text{argmin}_{(S,L)} \mathcal{L}(S-L, \Sigma^{(n)}_O) + \lambda_n (\gamma\|S\|_{1} + \|L\|_*)\]  s.t. $S-L \succ 0, L \succeq 0$
 		\STATE $\hat{E} \leftarrow \{(i,j) : i < j, \hat{S}_{ij} > T\}$
 			\\
         \STATE \textbf{Return:} $\hat{G} = (V, \hat{E})$
 	\end{algorithmic}
 \end{algorithm}

\paragraph*{Implementation}
Algorithm~\ref{alg:sl1} describes our latent structure learning method. 
We use the loss function 
from \citet{Wu17}:
\[\mathcal{L}(S-L, \Sigma^{(n)}_O) = \frac{1}{2} \tr((S-L)\Sigma^{(n)}_O(S-L)) - \tr(S-L).\]

We implement Algorithm~\ref{alg:sl1} using standard convex solvers. The recovered sparse matrix $\hat{S}$ does not have entries that are perfectly 0. Therefore, a key choice is to set a threshold $T$ to find the zeros in $\hat{S}$ such that

\begin{equation*}
	\tilde{S}_{ij} =
	\begin{cases}
	\hat{S}_{ij} & \text{if } \hat{S}_{ij} > T ,\\
	0 & \text{if } \hat{S}_{ij} \leq T. \\
	\end{cases}
  \end{equation*}

 We can then pass the nonzero entries of $\tilde{S}$ as dependencies to the generative model described in Section~\ref{sec:background}.

%% file: analysis.tex
Our goal is to provide guarantees on the probability that Algorithm~\ref{alg:sl1} successfully recovers the exact dependency structure.
The critical quantity in establishing these guarantees is $\| \Sigma_O^{(n)} - \Sigma_O \|$, the spectral norm of the estimation error of the covariance matrix.
We control this key quantity by characterizing the effective rank of the covariance matrix $\Sigma_O$ in Section~\ref{subsec:rank}.
We then introduce two different conditions on the effective rank; these enable us to derive our main result, consisting of improved sample complexities, in Section~\ref{subsec:formal}. 

We end our analysis by deriving an information-theoretic lower bound 
for the weak supervision structure learning problem in Section~\ref{subsec:lower}, characterizing the additional sample complexity cost versus the supervised setting. We show that although this cost can be unboundedly larger for the general latent setting, for the weak supervision case, it is reasonably small.

\subsection{Controlling the Covariance Estimation Error} 
\label{subsec:rank}
Structure learning algorithms for the supervised case \cite{Ravikumar11, Loh13} recover the structure with high probability given $\Omega(d^k \log m)$ samples, where $k \geq 2$ depends on the approach taken. The unsupervised (latent variable) algorithms in \citet{Chandrasekaran12, Wu17} require $\Omega(m)$ samples. 

The critical difference between these two classes of algorithms is found in their objectives. Note that the objective function for the algorithms of \citet{Ravikumar11, Loh13} contains the regularizer $\|\cdot\|_1$, while the algorithms in \citet{Chandrasekaran12, Wu17} instead have $\|\cdot\|_1 + \|\cdot\|_*$. The presence of the $\|\cdot\|_*$ norm in the objective for the latent settings is the key difference. Both classes of algorithms rely on the \emph{primal-dual witness} approach for their proofs of consistency. The dual norm of $\|\cdot\|_*$ is the spectral norm $\|\cdot\|$. As a result, a bound on $\|\Sigma_O^{(n)} - \Sigma_O\|$ (the estimated sample covariance matrix) is necessary---while a simpler entry-wise bound is sufficient for the supervised case. To ensure high-probability recovery, the unsupervised approaches rely on matrix concentration inequalities bounding $\|\Sigma_O^{(n)} - \Sigma_O\|$ that require $\Omega(m)$ samples. 

\paragraph*{Characterizing the Effective Rank}
To reduce this sampling rate, we leverage a refined measure of rank, the \emph{effective rank} \cite{Vershynin12}, defined as $$r_e(\Sigma_O) = \tr(\Sigma_O) / \|\Sigma_O\|.$$ The effective rank may be much smaller than the true rank; the notion that data matrices are approximately low-rank is well-known \cite{Udell18}. Characterizing the effective rank in the weak supervision setting enables us to apply sharper concentration inequalities. We use these tools to build on the analyses in \citet{Chandrasekaran12, Wu17} while providing improved rates. We note that \citet{Meng14} also considered the effective rank in a slightly different context. In the weak supervision setting, our characterization is tighter and we cover a wider range of cases.

Recall that the structure of $K_O^{-1}$ contains our key problem parameters---but $\Sigma_O$ does not. However, we show that 
$$r_e(\Sigma_O) \leq r_e(K_O^{-1}) + \frac{ \|v\|^2}{\|K_O^{-1}\|}.$$
Therefore, controlling the effective rank of $\Sigma_O$ can be done via the effective rank of $K_O^{-1}$. We can then characterize $r_e(\Sigma_O)$ in terms of structural information about the weak supervision sources. More details on this process are in the Appendix.  

\subsection{Conditions on the Effective Rank \& Main Results} 
\label{subsec:formal}

We provide two separate conditions on the effective rank, which lead to two different improved regimes for recovery in Algorithm~\ref{alg:sl1}. Let $0 < \tau \leq 1$ be a constant.

First, we define the source block decay (SBD) condition. In this case, the number of correlated clusters is $s = O(m^{\tau/2} / \log m)$ and $r_{e}(\Sigma_O) = O(m^\tau / \log m)$. Then we can recover the exact structure with probability at least $1-m^{-\tau}$ if the number of samples is
$n = \Omega( d^2 m^{\tau})$, where $d$ is the maximum dependency degree.

Next, we define the strong source block (SSB) condition, where $r_{e}(\Sigma_O) = O(d)$, with no requirement on $s$. Then, with probability at least $1 - m^{-\tau}$, we recover the exact dependency structure $G$ if
$n = \Omega( d^2 (1+\tau) \log m)$.

Note that the second sample complexity \emph{matches the supervised optimal rate} of $d^2 \log m$ samples. These conditions and the resulting sample complexities, along with those for the algorithms in \citet{bach2017learning} and \citet{Wu17} are summarized in Table~\ref{tab:table1}. Next, we explain them in more detail and provide the formal result.

\begin{table}[t]
  \begin{center}

    \begin{tabular}{@{}llll@{}}
      \toprule
       \textbf{Cond.} & \textbf{$r_e(\Sigma_O)$}      & \textbf{$s$}      & \textbf{Rate}    \\ \midrule
       Bach & $\text{none}$ & $\text{none}$ & $ \Omega( m\log m ) $       \\
       Wu & $ \text{none}$        & \text{none}                                                      & $ \Omega( d^2 m) $       \\
       SBD & $O(\frac{m^{\tau}}{ \log m})$ & $ O\left( \left(\frac{m}{\log^2 m} \right)^{\tau/(2-\tau)}\right) $ & $\Omega(d^2 m^{\tau})$ \\
      SSB & $O(d) $                       & $\text{none} $                                                   & $\Omega(d^2 \log m)$   \\ \bottomrule
      \end{tabular}      
      
  \end{center}
  \caption{Conditions and rates for latent variable structure learning in \citet{bach2017learning}, \citet{Wu17}, and Algorithm~\ref{alg:sl1} using the two conditions (SBD, SSB) we define. Shown are the conditions on the effective rank and the number of source clusters and the resulting sample complexities. Here $0 < \tau < 1$.}
    \label{tab:table1}
\end{table}

\begin{definition}[SBD Condition]
  \label{def:sbd}
The matrix $\Sigma_O$ satisfies the source block decay (SBD) condition if its effective rank $r_e(\Sigma_O)$ satisfies \[r_e(\Sigma_O) \leq  \frac{m^{\tau}}{(1+\tau) \log m}\] and the number of clusters $s$ satisfies \[s \leq \frac{m^{\frac{\tau}{2-\tau}}}{((1+\tau)\log m)^{2/(2-\tau)}}.\]
\end{definition}

This condition represents a mild assumption on the structure of $\Sigma_O$ (and, equivalently $K_O^{-1}$). 
It corresponds to mild eigenvalue decay in the source blocks, and a condition limiting the total number of blocks. 
In the weak supervision setting, this translates to the strength of some of the correlations in a cluster differing.
By exploiting this decay and controlling the total number of blocks $s$, we can obtain a sublinear sample complexity of $\Omega(d^2m^{\tau})$ for Algorithm~\ref{alg:sl1}.

\begin{definition}[SSB Condition]
  \label{def:ssb}
The matrix $\Sigma_O$ satisfies the strong source block (SSB) condition if its effective rank $r_e(\Sigma_O)$ satisfies $r_{e}(\Sigma_O) \leq c d,$ where $c$ is a constant.
\end{definition}

The second, alternate condition is equivalent to the presence of a cluster of sources that forms a strong voting block, dominating the other sources. With this condition, \emph{we can retrieve the optimal rate of} $\Omega(d^2 (1+\tau) \log m)$ \emph{from the supervised case}. We provide a more precise characterization for the effective rank bounds in the proof of the theorem in the Appendix. In particular, we show how to relate the effective rank of $\Sigma_O$ in terms of that of $K_O^{-1}$, enabling us to connect structural information like the quantities $d$ and $s$ to the conditions.

\paragraph{Additional Standard Conditions}
Next, we highlight the general conditions used by \citet{Chandrasekaran12} and \citet{Wu17} whose work we build on; we require these to hold in addition to the SBD or SSB conditions we define above.
Specifically, we use a series of standard quantities that control transversality, introduced by \citet{Chandrasekaran12} and \citet{Wu17}. Let $$h_X(Y) = \frac{1}{2}(XY + YX).$$ Let $\mathcal{P}_S$ denote orthogonal projection onto subspace $S$. The following terms are used to control the behavior of $h_X(\cdot)$ on the spaces $\Omega(S)$ and $T(L)$. For convenience, we simply use $\Omega$ and $T$ to denote these spaces. Let
\begin{align*}
  \alpha_{\Omega} &=   \min_{M \in \Omega, \|M\|_{\infty} = 1}    \|\p_{\Omega} h_{\Sigma_O}(M)\|_\infty , \\
  \delta_{\Omega} &=    \min_{M \in \Omega, \|M\|_{\infty} = 1}    \|\p_{\Omega^\perp} h_{\Sigma_O}(M)\|_\infty , \\
  \alpha_{T} &=    \min_{M \in T, \|M\| = 1}    \|\p_{T} h_{\Sigma_O}(M)\| , \\
  \delta_{T} &=    \min_{M \in T, \|M\| = 1}     \|\p_{T^{\perp}} h_{\Sigma_O}(M)\| , \\
  \beta_{T} &=    \max_{M \in T, \|M\|_{\infty} = 1}    \|h_{\Sigma_O}(M)\|_\infty , \\  
  \beta_{\Omega} &=    \max_{M \in \Omega, \|M\| = 1}     \|h_{\Sigma_O}(M)\| .
\end{align*}
We set 
\[\alpha = \min\{\alpha_\Omega, \alpha_T \}, \text{   } \beta = \max\{ \beta_T, \beta_\Omega \}, \text{   } \delta = \max\{\delta_\Omega, \delta_T\}.\]

The following irrepresentability conditions are inherited from \citet{Wu17} and are generalizations of standard conditions from the graphical model literature~\cite{Ravikumar11,zhang2014sparse}: there exists $\nu \in (0,1/2)$ with $$\delta/\alpha < 1 - 2\nu, \qquad \text{ and }$$ $$\mu(\Omega) \xi(T) \leq \frac{1}{2} \left( \frac{\nu \alpha}{(2-\nu)\beta} \right)^2.$$ 

Finally, let $\psi_1$ be the largest eigenvalue of $\Sigma_O$, $\psi_m$ be the smallest, let $K_{O,\min}$ be the smallest non-zero entry in $K_O$, and $\sigma$ be the nonzero eigenvalue of $zz^T$. We set \[\gamma = \frac{\nu \alpha}{2 d \beta (2 - \nu)}.\]

\paragraph*{Main Results}
We now present the formal result for the consistency of Algorithm~\ref{alg:sl1}. First, the source block decay case:

\begin{theorem}[Source Block Decay Case]
Let $0 < \tau \leq 1$ be a constant. Suppose that the standard conditions above and the SBD condition are met. Set 

\[\lambda_n = \max \{1, \gamma^{-1}\} \frac{(3-2\nu) c_1 \psi_1 \sqrt{m^\tau}}{ \psi_m \sqrt{n}}.\] 

Let 
\[\rho_1 = \left[ \frac{6c_2\beta(3-2\nu)(2-\nu) \psi_1}{\nu \alpha^2 \psi_m} \max\{\frac{\gamma}{K_{O,\min }}, \sigma^{-1}, \frac{1}{\psi_m}\}\right]^2.\]
If the number of samples $n$ satisfies 
\[n > \rho_1 d^2 m^{\tau},\]
and we run Algorithm~\ref{alg:sl1}, then, with probability at least $1 - m^{-\tau}$, we recover the exact structure $G$.
\label{thm:mainrate}
\end{theorem}
\addtocounter{theorem}{-1}

Next, the strong source block case:
\begin{theorem}[Strong Source Block Case]
Suppose instead that in addition to the standard conditions, the SSB condition holds. 
Set
\[\lambda_n = \max \{1, \gamma^{-1}\} \frac{(3-2\nu)c_4 c_{2} \psi_1 d (1 + \tau) \log(m)}{ \psi_m n}.\]

Let 
\[\rho_2 =  \frac{6\beta c_2c_4 (3-2\nu)(2-\nu) \psi_1}{\nu \alpha^2 \psi_m } \max \left\{ \frac{\gamma}{K_{O,\min}}, \sigma^{-1}, \frac{1}{\psi_m} \right\}.\]

If the number of samples $n$ satisfies 
\[ n > \rho_2 (1+\tau) d^2 \log(m),\]
then, with probability at least $1 - m^{-\tau}$, we recover the exact structure $G$.
\end{theorem}

We provide a formal proof of Theorem~\ref{thm:mainrate} in the Appendix. The proof modifies the proof technique in \citet{Wu17} by applying stronger concentration inequalities and adapting the resulting analysis.

\section{Information-Theoretic Lower Bound}
\label{subsec:lower}

So far, we have worked with a particular algorithm, showing that under the stronger of our two conditions, the sample complexity matches the optimal one of $\Omega(d^2 \log m)$ for supervised structure learning. Now we explore the general question of the fundamental limits of structure learning with latent variables. This is accomplished by deriving information-theoretic lower bounds on sample complexity: bounds that show that \emph{for any} algorithm, at least a certain number of samples is required to avoid incurring a particular probability of error. 

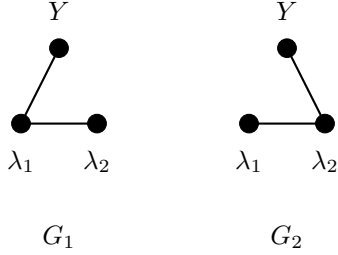
\begin{figure}[t]
	\begin{center}
		\begin{tikzpicture}[scale=0.5]
		\draw[fill=black] ( 0, 0 ) circle (7pt);
		\draw[fill=black] ( 1, 2 ) circle (7pt);
		\draw[fill=black] ( 2, 0 ) circle (7pt);
	    \draw[thick] (0,0)  -- (1,2);
	    \draw[thick] (2,0)  -- (0,0);
	    \node at (0, -1.0) {$\lf_1$};
	    \node at (2, -1.0) {$\lf_2$};
	    \node at (1, 3.0) {$Y$};
		\node at (1, -3.0) {$G_1$};
		
		\draw[fill=black] ( 6, 0 ) circle (7pt);
		\draw[fill=black] ( 7, 2 ) circle (7pt);
		\draw[fill=black] ( 8, 0 ) circle (7pt);
		\draw[thick] (8,0)  -- (7,2);
		\draw[thick] (8,0)  -- (6,0);
		\node at (6, -1.0) {$\lf_1$};
		\node at (8, -1.0) {$\lf_2$};
		\node at (7, 3.0) {$Y$};
		\node at (7, -3.0) {$G_2$};
		\end{tikzpicture}
	\end{center}
	\caption{Simple example for latent variable structure learning. In these two graphs, $Y$ is not connected to every source. Observing $\lf_1$ and $\lf_2$ does not allow us to establish which of $\{G_1, G_2\}$ is the true model, regardless of the number of samples.}
		\label{fig:small_example}
\end{figure}

First, we may ask what happens in the general latent-variable case. In this setting, we do not need to have $Y$ connected to each of the $\lf_i$ source variables; $Y$ may be connected to just some of these sources. Even if we ensure that the class of graphs we are working over is connected overall, there are graphs that cannot be distinguished, with any number of samples. One such example is shown in Figure~\ref{fig:small_example}. Here, we have two graphs, $G_1$ and $G_2$, where the only difference is that in one case, there is an edge between $Y$ and $\lf_1$, while in the other, there is an edge between $Y$ and $\lf_2$. By observing only $\lf_1$ and $\lf_2$, but not $Y$, we cannot distinguish between these two graphs.

For this reason, working in the fully-general latent structure learning setting leads to uninteresting results. Instead, we again work in the weak supervision setting where $Y$ is connected to all of the $\lf_i$'s. We already know, from our algorithmic analysis, that in certain cases we can recover the structure with $\Omega(d^2 \log m)$ samples, and this quantity is optimal even in the supervised case. Certainly we expect that the presence of the latent variable $Y$ will require more samples (in terms of lower bounds). In Theorem~\ref{thm:lb_unsupervised} we quantify this difference.

The strategy used to derive information-theoretic lower bounds is to construct a collection of graphs along with a set of parameters and to use Fano's inequality (or related methods) that rely on a notion of distance between pairs of graphs in the collection. The smaller this distance, the larger the number of samples required to distinguish between a pair of graphs. Our approach is to consider a collection of graphs used to derive the $\Omega(d^2 \log m)$ lower bound, and to construct the equivalent collection in the latent-variable weak supervision case. We then compute how much larger the number of samples required for reliably selecting the correct graph is for the unsupervised versus supervised case.

Let $\mathcal{G}_{\text{ws}}$ be the class of graphs on $m+1$ nodes ($m$ sources and 1 latent node connected to all of the other nodes) with maximum degree $d$, structured according to our exponential family model, restricted to the setting where only edge parameters are non-zero, and all such edge parameters are $\theta$. Let $M = |\mathcal{G}_{\text{ws}}|$. Our main result is 

\begin{theorem}
Any decoding procedure to determine $G$ from samples of $\lf_1, \ldots ,\lf_m$ will have maximum probability of error at least $\delta - \frac{1}{\log M}$ if the number of samples $n$ is upper-bounded as
\[ n  < (1-\delta) \frac{\log (m(m-1)/2)}{2 \theta (1 - 4(\exp(4\theta)+3)^{-1} - \tanh^2(\theta))}.\]
\label{thm:lb_unsupervised}
\end{theorem}

As expected, that the number of samples here is larger than supervised version, where the expression simply has a $2 \theta \tanh \theta$ in the denominator~\cite{Santhanam12}. In particular, the number of additional samples $n_{\Delta}$ we need is given by 
\begin{align*}
n_{\Delta} = &\frac{(1-\delta) \log (m(m-1)/2)}{2\theta} \times \\
 &\left[ \frac{1}{1 - 4(\exp(4\theta)+3)^{-1} - \tanh^2(\theta)} - \frac{1}{\tanh \theta} \right].
\end{align*}

This quantity characterizes the cost in sample complexity due to the weak supervision setting. We observe, however, that in the limit of $\theta \rightarrow 0$, this relative cost is not too high. This is the regime of interest for $d, m \rightarrow \infty$, where we end up requiring that $\theta \rightarrow 0$ in order to avoid an exponential sample complexity \cite{Santhanam12}. Then, the \emph{relative} version of the cost above can be seen, after some algebra, to be upper bounded by 2. That is, we need no more than twice as many samples as in the unsupervised case to avoid an unreliable encoder.

We can interpret this result in the following way. As we intuitively expect, latent variable structure learning requires more samples than the fully-supervised version; potentially, infinitely many samples are required. However, the weak-supervision setting provides us with a tractable scenario, where the lower bounds are not much larger than the supervised equivalents.

We briefly comment on the approach to Theorem~\ref{thm:lb_unsupervised}. The collection considered here is made by taking the graphs where all of the $\lf_i$'s have no edge between them and adding a single edge between $\lf_s$ and $\lf_t$. Thus there are $\binom{m}{2}$ such graphs in the collection. In the supervised setting, there is just one such edge per graph; in our latent-variable setting, there are $m$ additional edges, those between each $\lf_i$ and $Y$. Intuitively, the challenge when distinguishing between graphs is to ascertain whether a pair of nodes are connected by an edge. In the latent version, this task is harder since all pairs of nodes are additionally connected through $Y$.

%% file: exp.tex
We evaluate our structure learning method on real-world applications ranging from medical image classification to relation extraction over text.
We compare our performance to several common weak supervision baselines: an unweighted majority vote of the weak supervision source labels, a generative modeling approach that assumes independence among weak supervision sources~\cite{Ratner16}, and a generative model using dependency structure learned with an existing structure learning approach for weak supervision~\cite{bach2017learning}. We compare these baselines to the same generative modeling technique using dependencies learned by our approach and report performance of the discriminative model trained on labels from the generative model. 

\begin{table*}[t]
	\centering    
	\begin{tabular}{@{}lccccccccc@{}}
		\toprule
		& \multicolumn{1}{l}{} & \multicolumn{1}{l}{} & \multicolumn{1}{l}{} & \multicolumn{1}{l}{} & \multicolumn{1}{l}{} & \multicolumn{1}{l}{}  & \multicolumn{2}{c}{\textbf{Improvement Over}} \\ \cmidrule(l){8-9} 
		\textbf{Application} & \textbf{$m$}      & \textbf{($s,d$)}  & \textbf{MV}          & \textbf{Indep.}      & \textbf{Bach et al.}     & \textbf{Ours}        & \textbf{Indep.}    & \textbf{Bach et al.}    \\ \midrule
		Bone Tumor           & 17                    & (2,3)                    & 65.72                & 67.32                & 67.83              & 71.96                & +4.64                   & +4.13               \\
		CDR                  & 33                   & (22,14)                  & 47.74                & 54.60                & 55.90               & 56.81                & +2.21                    & +0.91              \\
		IMDb                 & 5                    & (1,4)                    & 55.21                & 58.80                & 60.23               & 62.71                & +3.91                   &  +2.48               \\
		MS-COCO              & 3                    & (1,2)                    & 57.95                & 59.47                & 59.47               & 63.88                & +4.41                   &  +4.41              \\ \bottomrule
	\end{tabular}
	\caption{Statistics for weak supervision tasks ($m$: number sources, $s$: number of cliques, $d$: max. degree of source). F1 scores of discriminative models trained on labels generated by majority vote (MV), a generative model with no dependencies (Indep.), a generative model with dependencies learned by a prior structure learning approach for weak supervision (Bach et al), and by our approach (Ours).}
	\label{table:results}
\end{table*}

The weak supervision sources for these tasks include a variety of signals exploited in prior work, such as user-defined heuristics, distant supervision from dictionaries, and regular expression patterns.
Recovering the dependencies among these sources using our approach leads to an improvement of up to $4.64$ F1 points over assuming independence, and up to $4.41$ F1 points over using the dependencies learned by an existing structure learning method.
Finally, we run simulations over synthetic data to explore our performance compared to existing methods under the two conditions of strong source block (SSB) and source block decay (SBD).

\subsection{Real-World Tasks}
We use a generative model over the labels from the weak supervision sources to generate probabilistic training labels and then train a generic discriminative model associated with the task~\cite{Ratner16,bach2017learning,varma2017inferring}. We report the test set performance of the same discriminative model, trained on labels from the different baselines and our approach in Table~\ref{table:results}.

\paragraph{Task Descriptions}
We describe the different weak supervision tasks, the associated weak supervision sources, and the discriminative model used to perform classification. The \textbf{Bone Tumor} task is to classify tumors in X-rays as aggressive or non-aggressive~\cite{varma2017inferring}. The discriminative model is a logistic regression model over hundreds of shape, texture, and intensity-based image features. The weak supervision sources are a combination of user-defined heuristics and decision trees over a different set of features extracted from the X-rays. 

The \textbf{CDR} task is to detect relations among chemicals and disease mentions in PubMed abstracts~\cite{bach2017learning,cdr}. The discriminative model is an LSTM~\cite{graves2005framewise} that takes as input sentences containing the mentions. The weak supervision sources are a combination of distant supervision from the Comparative Toxicogenomics Database~\cite{davis2016comparative} and user-defined heuristics. 

The \textbf{IMDb} task is to classify plot summaries as describing action or romantic movies~\cite{varma2017flipper}. The discriminative model is an LSTM that takes as input the entire plot summary. The weak supervision sources are user-defined heuristics that look for mentions of specific words in the plot summary.
The \textbf{MS-COCO} task is to classify images as containing a person or not~\cite{varma2017socratic}. The discriminative model is GoogLeNet. The weak supervision sources are user-defined heuristics written over the captions associated with the images.

\paragraph{Performance}
Our method learns dependencies among the supervision sources for each of the tasks described above, which leads to an average improvement of $3.80$ F1 points over the model that assumes independence. We also compare to a prior structure learning approach for weak supervision~\citet{bach2017learning}. For the MS-COCO task, \citet{bach2017learning} is unable to learn \emph{any} dependencies while our method learns a single pairwise dependency, which improves performance by $4.41$ F1 points. For the Bone Tumor task, our method identifies $2$ cliques with $3$ supervision sources. The first clique consists of heuristics that all rely on features related to edge sharpness along the lesion contour of the tumor, while the sources in the second clique rely on features describing the morphology of the tumor. Incorporating these dependencies in the generative model improves over \citet{bach2017learning} by $4.13$ F1 points. Finally, for the IMDb task, our method learns a clique involving $4$ sources while \citet{bach2017learning} only learns $3$ pairwise dependencies among the same sources. Learning a clique improves performance by $2.48$ F1 points. 

\subsection{Simulations}
\begin{figure}[t]
	\centering
	\includegraphics[width=0.75\columnwidth]{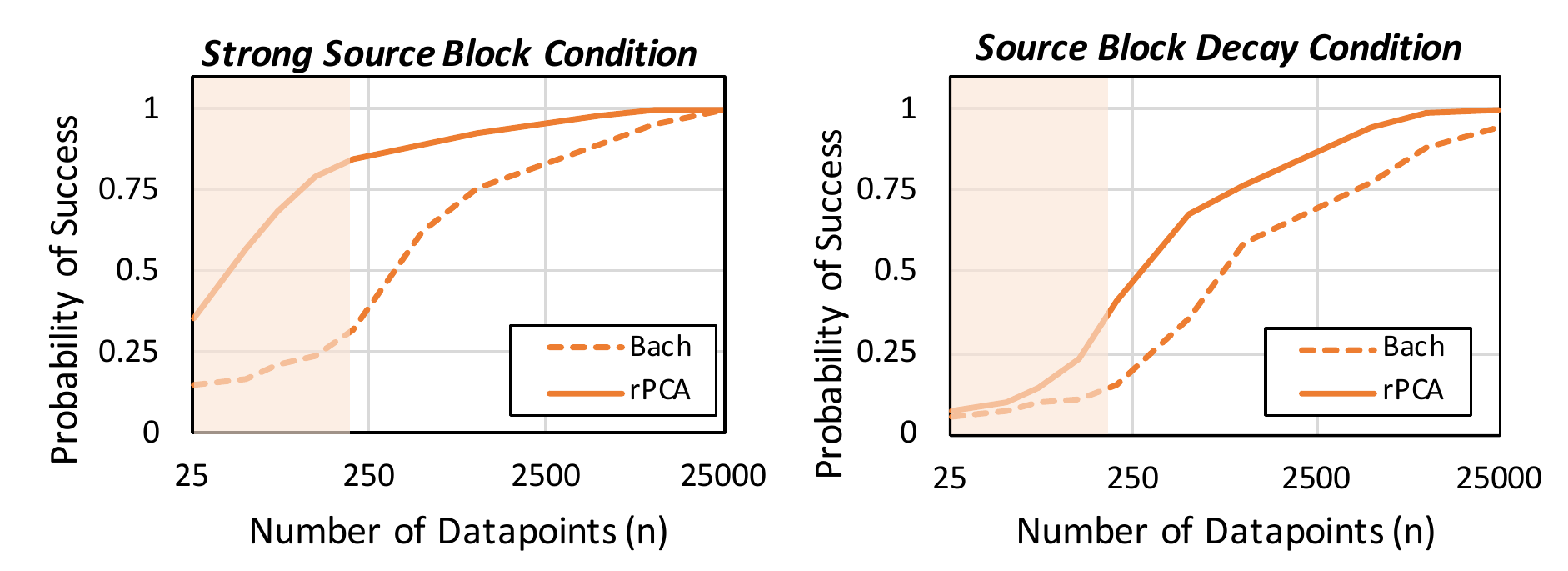}
	\caption{Shaded region shows where $n < m$. With the strong source block condition, our method significantly outperforms existing method. Without this condition, neither methods work well when $n<m$, as expected.}
	\label{fig:sim}
\end{figure}

We also perform simulations over synthetic data using $200$ weak supervision sources to explore how our performance compares to \citet{bach2017learning} under the two conditions on effective rank described in Section~\ref{sec:analysis}, the strong source block (SSB) condition and the source block decay (SBD) condition. We define success as how often these methods are able to learn the true dependencies and plot our results in Figure~\ref{fig:sim}. We first generate labels from supervision sources to match the strong source block condition by ensuring there exists a single cluster of strongly correlated sources along with other more weakly correlated sources. We observe that our method performs significantly better than \citet{bach2017learning}, and is capable of recovery in the regime where $n$ starts at roughly $\log m$ and goes up to roughly $m$. Second, we simulate the source block decay condition by generating multiple cliques of sources where a single dependency in each clique is stronger than the rest. We continue to perform better compared to \citet{bach2017learning} under this condition and across all values of $n$. 

%% file: conc.tex
The dependency structure of generative models significantly affects the quality of the generated labels. However, learning this structure without any ground truth labels is challenging. We present a structure learning method that relies on robust principal component analysis to estimate the dependencies among the different weak supervision sources. We prove that the amount of unlabeled data required to estimate the true structure can scale sublinearly or even logarithmically with the number of weak supervision sources,
improving over the standard sample complexity, which is linear. Under certain conditions, we match the information-theoretic optimal lower bound in the supervised case. 
Empirically, this translates to our method outperforming traditional structure learning approaches by up to $4.41$ F1 points and methods that assume independence by up to $4.64$ F1 points.

%% file: glossary.tex
The glossary can be found in Table~\ref{table:glossary} below.

\begin{table*}[h]
\centering
\begin{tabular}{l l}
\toprule
Symbol & Used for \\
\midrule
$\x$ & Data point, $\x \in \mathcal{X}$ \\
$n$ & Number of data points \\
$\y$ & Latent label  \\
$\lf_i$ & Weak supervision sources output by the $i$th source for $\x$ \\
$m$ & Number of sources \\
$G$ & Source dependency graph, $G = (V,E)$, $V=\{\lf_1,...,\lf_m\} \cup \{\y\}$ \\
$f_G$ & Density of weak supervision sources $\lambda_1 ... \lambda_m$ and latent variable $\y$\\
$d$ & Maximum degree of weak supervision sources in $G$ \\
$s$ & Number of cliques of dependent weak supervision sourcess in $G$ \\
$O$ & The set of observable variables, i.e., weak supervision sources (but not the label $\y$) \\
$ \mathcal{S}$ & The set of unobserved variables, i.e., the latent label $\y$ \\ 
$\Sigma$ & Covariance matrix of $O \cup \mathcal{S}$, $\Sigma = \Cov{}{ O \cup \mathcal{S} }$ \\
$K$ & The inverse covariance matrix $K = \Sigma^{-1}$ \\
$\Sigma_O$ & Covariance matrix of $O$, the observed variables. Neither $\Sigma_O$ nor $\Sigma_O^{-1}$ are graph structured \\
$K_O$ & Sub-block of inverse covariance matrix $K$ corresponding to observed variables \\
$zz^T$ & Low rank matrix that encodes the parameters of the graph $f_G$ such that $K_O  = \Sigma_O^{-1} + zz^T$\\
$T(L)$ & Tangent space for the low-rank component in robust PCA, $T(L) = \{UX^T + YV^T \text{ } | \text{ } Y_1, Y_2 \in \mathbb{R}^{m \times r} \}$ \\
$\Omega(S)$ & Tangent space for the sparse component, $\Omega(S) = \{X \in \mathbb{R}^{m \times m} \text{ }  |  \text{ }  \supp(N) \subseteq \supp(S)\}$\\
$ \xi(T(L))$ & Measurement of diffuseness of the low-rank term, $ \xi(T(L)) = \max_{N \in T(L), \|N\| \leq 1} \|N\|_{\infty}$ \\
$\mu(\Omega(S))$ & Measurement of sparsity, $\mu(\Omega(S)) = \max_{N \in \Omega(S), \|N\|_{\infty} = 1} \|N\|$\\
$\tau$ &  Constant between 0 and 1 that controls the sampling rate and the error probability \\
$\psi_1$ & Smallest eigenvalue of $\Sigma_O$ \\
$\psi_m$ & Largest eigenvalue of $\Sigma_O$  \\
$\gamma$ &  Hyperparameter in Algorithm~\ref{alg:sl1}\\
$\lambda_n$ & Positive eigenvalue of $L = zz^T$\\
$\sigma$ &  Constant related to $\lambda_n$ that controls sample complexity of Algorithm\\
$K_{O,min}$ & Smallest non-zero entry of $|K_O|$ \\
$\alpha_{\Omega}$ &   $\min_{M \in \Omega, \|M\|_{\infty} = 1}    \|\p_{\Omega} h_{\Sigma_O}(M)\|_\infty$ \\
$\delta_{\Omega}$ &    $\min_{M \in \Omega, \|M\|_{\infty} = 1}    \|\p_{\Omega^\perp} h_{\Sigma_O}(M)\|_\infty$ \\
$\alpha_{T}$ &    $\min_{M \in T, \|M\| = 1}    \|\p_{T} h_{\Sigma_O}(M)\|$ \\
$\delta_{T}$ &    $\min_{M \in T, \|M\| = 1}     \|\p_{T^{\perp}} h_{\Sigma_O}(M)\|$ \\
$\beta_{T}$ &    $\max_{M \in T, \|M\|_{\infty} = 1}    \|h_{\Sigma_O}(M)\|_\infty$ \\  
$\beta_{\Omega}$ &    $\max_{M \in \Omega, \|M\| = 1}     \|h_{\Sigma_O}(M)\|$ \\
$\alpha$  & $\min\{\alpha_\Omega, \alpha_T \}$ \\
$\beta $  & $\max\{ \beta_T, \beta_\Omega \} $\\
$\delta$  & $\max\{\delta_\Omega, \delta_T\} $\\

\end{tabular}
\caption{
	Glossary of variables and symbols used in this paper.
}
\label{table:glossary}
\end{table*}

%% file: related.tex
A common alternative to hand labeling data is using weak supervision sources, such as distant supervision~\cite{craven1999constructing,mintz2009distant}, multi-instance learning~\cite{riedel2010modeling,hoffmann2011knowledge} and heuristics~\cite{bunescu2007learning,shin2015incremental}. Estimating the accuracies of weak supervision sources without ground truth labels is a classic problem~\cite{dawid1979maximum}. Methods like crowdsourcing~\cite{dalvi2013aggregating,joglekar2015comprehensive,zhang2016spectral}, and boosting~\cite{schapire2012boosting} are common approaches; however, we focus on the case in which \emph{no labeled data} is required. 

Recently, generative models have been used to combine various sources of weak supervision~\cite{alfonseca2012pattern,takamatsu2012reducing,roth2013combining,ratner2016data,Ratner19}. Most previous work assumes that the structure of these models is user-specified. \citet{bach2017learning} recently showed that it is possible to learn dependencies with a sample complexity that scales quasilinearly with the number of sources. \citet{varma2017inferring} inferred dependencies using the code used to define the weak supervision sources. Our method improves over~\citet{bach2017learning} by reducing the dependence on the number of sources to sublinear, and, under stronger conditions, logarithmic, and is able to learn dependencies that are not explicit in the code, thus improving over~\citet{varma2017inferring} as shown in Section~\ref{sec:exp}.

Structure learning outside the context of weak supervision can be roughly divided into the supervised and unsupervised case, which require access to ground truth labels and not, respectively. Within these, we can further split the methods into node-wise and matrix-wise methods. Node-wise methods, like~\citet{bach2017learning}, use regression on a particular node to recover that node's neighborhood~\cite{candes2007dantzig,ravikumar2010high,wainwright2008graphical,tibshirani1996regression} and matrix-wise methods like ours use the inverse covariance matrix to determine the structure~\cite{Loh13,Chandrasekaran12}. 
The canonical matrix-wise method in the supervised case is the graphical Lasso algorithm (GLASSO)~\cite{Friedman08}. 

Analysis of the graphical lasso applied to sparse inverse covariance matrices was studied in \citet{Ravikumar11}, which achieves a sample complexity of $d^2 \log m$. 	The key question is then when the inverse covariance (precision) matrix of a random vector is sparse. In the classical, Gaussian case, the sparsity is governed by the graphical model associated with the vector: if a pair of variables are independent conditioned on all the other variables (i.e., there is no edge between the associated nodes in their graph), the precision matrix is 0 at the corresponding entry. This is not necessarily the case for non-Gaussian models. For discrete models, \cite{Loh13} characterizes the situations when the precision matrix is indeed graph-structured. In many cases, it is necessary to form a larger, \emph{generalized} covariance matrix. However, for graphs with singleton separator sets, the normal precision matrix is graph-structured.

The idea of using robust PCA for separating the sparse part of the precision matrix (encoding the graph structure) from the low-rank matrix that captures the marginalizing effects dates back to the original papers on robust PCA \cite{Chandrasekaran12}. For Gaussian graphical models with latent variables, \citet{Chandrasekaran11} produced the seminal work \citet{Chandrasekaran12}. More recent work follows the same approach, but modifies the loss function \cite{Wu17} and relaxes the Gaussian assumptions. On the other hand, \citet{Wu17}	 still requires a partially Gaussian model in order to induce sparsity, while our work operates in the discrete case entirely by leveraging the singleton separator set criteria. Both of these works lead to a $\Omega(m)$ rate. The work \citet{Meng14} is closer to the spirit of our approach; one of their results also considers using the effective rank by applying a theorem from \citet{Lounici14}. However, our work and theirs has key differences: they consider the Gaussian rather than discrete setting, they are interested in model estimation rather than selection. Finally, they work in a general setting; our work in the weak supervision setting	enables us to more tightly characterize the effective rank in terms of key sparsity parameters, while they leave the effective rank as a parameter that can only be measured.

The major work for structure learning in the weak supervision regime is \citet{bach2017learning}. This is a fundamentally different approach, building on the node-wise methods and using a pseudo-likelihood to estimating the structure. The key requirement of \citet{bach2017learning} is the maximum number of dependencies $d$. However, this maximum is taken over \emph{all variables}, both observed and latent. In the weak supervision scenario, there is a dependency between each weak supervision sources and the latent true label, and therefore this degree $d$ always takes the value $m$ (Corollary 2 in \citet{bach2017learning}), leading to a rate of $\Omega(m \log m)$. The key advantage of our work is that for us, $d$ is taken over the observed variables only---and therefore can be much smaller than $m$. This enables us to obtain sublinear (and even logarithmic) rates in $m$, and to better scale with the sparsity of the model. 

%% file: proofs.tex
Next, we give proofs of our results, starting with Lemma~\ref{lem:ab}.

\begin{proof}
Our goal is to bound the product $\mu(K_O) \xi(zz^T)$. Bounding $\mu(K_O)$ is easy: we apply the simple bound $\mu(K_O) \leq d$ (Proposition 3 in \citet{Chandrasekaran11}). 

We must bound the $\xi(zz^T)$ term, which we do as follows. First, since $zz^T$ is symmetric, it has the same row-space and column-space. Let this row-space be denoted $\text{rs}(zz^T)$. Define $\beta(S) := \max_i \|P_S e_i\|_2$, where $P_S$ is projection onto the subspace $S$ and $e_i$ is the $i$th standard basis vector. Then, from Proposition 4 in \citet{Chandrasekaran11}), \[\xi(zz^T) \leq 2 \beta(\text{rs}(zz^T)).\] 

Since $zz^T$ is rank-one, its row-space is simply spanned by $z$ and $\beta(\text{rs}(zz^T)) = \|\bar{z}\|_{\infty}$, where $\bar{z}$ is $z / \|z\|$. Now, applying the definition of $z$,
\begin{align}
\beta(\text{rs}(zz^T)) = \|\bar{z}\|_{\infty} = \frac{\|z\|_{\infty}}{\|z\|} = \frac{\| \Sigma_{O}^{-1} \Sigma_{OS}\|_{\infty}}{\| \Sigma_{O}^{-1} \Sigma_{OS}\|} .
\label{eq:betabd}
\end{align}

Now, we can upper bound the numerator
\[\|\Sigma_{O}^{-1} \Sigma_{OS}\|_{\infty} \leq \| \Sigma_{O}^{-1} \|_{\infty} \|\Sigma_{OS}\|_{\infty}.\]

We lower bound the denominator as
\[\| \Sigma_{O}^{-1} \Sigma_{OS}\| \geq \sigma_{\min}(\Sigma_O^{-1}) \|\Sigma_{OS}\| = \frac{ \|\Sigma_{OS}\|}{\sigma_{\max}(\Sigma_O)} .\]

Using these bounds in \eqref{eq:betabd}, we have that 
\begin{align}
\beta(\text{rs}(zz^T))  \leq \left(  \frac{ \sigma_{\max}(\Sigma_O) \| \Sigma_{O}^{-1} \|_{\infty} \|\Sigma_{OS}\|_{\infty} }{\|\Sigma_{OS}\|} \right).
\label{eq:betabd2}
\end{align}

Recall that $a_{\min}, a_{\max}$ are the smallest and largest terms in $\Sigma_{OS}$, respectively.  Similarly, recall that $c_{\min}, c_{\max}$ are the smallest and largest terms in $\Sigma_O$. We have that $\|\Sigma_{OS}\|_{\infty} = a_{\max}$. Also, $\|\Sigma_{OS}\| \geq \sqrt{m}a_{\min}$, so that 
\[\frac{\|\Sigma_{OS}\|_{\infty}}{\|\Sigma_{OS}\|} \leq \frac{a_{\max}}{\sqrt{m} a_{\min}}.\]

Bounding $\|\Sigma_O^{-1}\|_{\infty}$ is slightly more challenging. Recall the definition of a symmetric diagonally dominant (SDD) matrix. A matrix $J \in \mathbb{R}^{m \times m}$ is SDD if it is symmetric and if $\Delta_i(J) := |J_{ii}| - \sum_{j \neq i} |J_{ij}| \geq 0$ for all $i = 1, \ldots, m$. It is often the case that covariance matrices are SDD (for example, this is the case for the covariances of Gaussian free field models). Even if $\Sigma_O$ is not SDD, we can make it SDD by performing the operation $\Sigma_O \leftarrow \Sigma_O + \nu I$, for some $\nu$ satisfying $\nu \leq (m-1) c_{\max}$. This operation is equivalent to adding independent noise with variance $\nu$ to each of the labeling functions. Critically, this does not affect the off-diagonal entries of $\Sigma$, which are what we wish to recover.

Thus we take $\Sigma_O \leftarrow \Sigma_O + \nu I$ so that $\Sigma_O$ is SDD. Then we apply the following tight bound on the $\infty$ norm of the inverse of a SDD matrix \cite{Hillar14}:
\begin{align*}
\| \Sigma_O^{-1}\|_\infty \leq \frac{3m-4}{2c_{\min} (m-2)(m-1)} \leq \frac{8}{5c_{\min} m}.
\end{align*} 

Finally, we must bound the largest singular value of $\Sigma_O$. Here, we use a Gerschgorin-style bound \cite{Qi84}: $\sigma_{\max}(\Sigma_O)$ is at most the largest row or column sum (excluding diagonal elements) plus the largest diagonal element. For us, $\sigma_{\max}(\Sigma_O) \leq \nu + m c_{\max} \leq (2m-1) c_{\max}$. 

Putting these results into \eqref{eq:betabd2}, we obtain
\begin{align*}
\beta(\text{rs}(zz^T))  \leq  &\frac{8(2m-1)c_{\max} a_{\max}}{5m^{3/2} c_{\min} a_{\min}}   \\
&\leq \frac{3.2}{\sqrt{m}} \frac{c_{\max} a_{\max}}{c_{\min} a_{\min}}.
\end{align*}

Thus,
\[ \xi(zz^T) \leq \frac{6.4}{\sqrt{m}} \left(\frac{c_{\max}}{c_{\min}}\right) \left(\frac{a_{\max}}{a_{\min}}\right).\]
Multiplying by $\mu(K_O) \leq d$ gives the result.
\end{proof}

\paragraph{Proof of Theorem~\ref{thm:mainrate}}

Now we prove Theorem~\ref{thm:mainrate}. 
\begin{proof}
Our approach proceeds in two steps. First, we bound the \emph{effective rank} \cite{Vershynin12} of our estimate of $\Sigma_O$, the covariance matrix of the observed sources. Next, we apply a pair of concentration bounds for estimating $\Sigma_O$. Afterwards, we show how to adapt the proof of Theorem 4.1 in \citet{Wu17} to obtain the result in Theorem~\ref{thm:mainrate}.

\paragraph{Effective Rank} The effective rank of a matrix $A$ is
\[r_e(A) = \frac{\tr(A)}{\|A\|}.\]
This quantity can be far smaller than the actual rank. As we shall see, sharp concentration bounds for estimating $\Sigma_O$ can be derived by exploiting $r_e(\Sigma_O)$. We begin by bounding this quantity in our setting. Applying the matrix inversion lemma, we have that
\begin{align*}
\Sigma_O &= K_O^{-1} + (K_S - K_{OS}^T K_O^{-1} K_{OS})^{-1} K_O^{-1}K_{OS}^T (K_O^{-1}K_{OS}^T)^T \\
&= K_O^{-1} + v v^T,
\end{align*}
where $v =  (K_S - K_{OS}^T K_O^{-1} K_{OS})^{-\frac{1}{2}} K_O^{-1}K_{OS}^T$. Then,
\begin{align*}
r_e(\Sigma_O) &= \frac{\tr(\Sigma_O)}{\|\Sigma_O\|} \\
&= \frac{\tr( K_O^{-1} + vv^T)}{\|\Sigma_O\|} \\
&= \frac{\tr(K_O^{-1}) + \tr(vv^T)}{\|\Sigma_O\|} \\
&=  (\tr(K_O^{-1} + \tr(vv^T))(\lambda_{\min}(\Sigma_O^{-1})) \\
&= (\tr(K_O^{-1})+\tr(vv^T))(\lambda_{\min}(K_O - zz^T))  \\
&\leq (\tr(K_O^{-1})+ \tr(vv^T))(\lambda_{\min}(K_O) + \lambda_{\max}(-zz^T)) \\
&=  \frac{\tr(K_O^{-1}) + \|v\|^2}{\|K_O^{-1}\|} \\
&=  r_e(K_O^{-1})  + \frac{ \|v\|^2}{\|K_O^{-1}\|}.
\end{align*}
Here, we upper bounded the effective rank of $\Sigma_O$ in terms of the effective rank of $K_O^{-1}$. The motivation for doing so is that $K_O^{-1}$ is more tractable to analyze with respect to our key quantities, such as $d$ and $s$. Recall that $K_O$ is sparse matrix. Moreover, it is (a permutation) of a block diagonal matrix. Then, $K_O^{-1}$ is also block diagonal and sparse. 

Next, we motivate the two conditions from Theorem~\ref{thm:mainrate}. Recall that $s$ is the number of cliques among our labeling functions.  Let $C_1, C_2, \ldots, C_s$ be the cliques that correspond to the variables $\lf_1, \ldots, \lf_m$, with $\sum_{j=1}^s |C_j| = m$ and $|C_j| \leq d$ for all $1 \leq j \leq s$. With slightly abuse of notation, we also refer to $C_j$ as the corresponding submatrix in $K_O^{-1}$.

For our first condition, note that in general $\tr(C_i) \leq |C_i| \lambda_{\max}(K_O^{-1})$. We assume that $\tr(C_i) \leq \frac{1}{2} |C_i|^{\tau/2} \lambda_{\max}(K_O^{-1})  \leq \lambda_{\max}(K_O^{-1}) |C_i|^{\tau/2}$. Effectively, we are assuming eigenvalue decay in each clique of sources with rate $\tau/2$; this is reasonable, since these blocks behave like the adjacency graph of a complete graph. The largest eigenvalue of such an adjacency matrix is large, but all remaining eigenvalues are small. Now, under this assumption, we have, by Holder's inequality,
\[r_e(K_O^{-1}) = \frac{\sum_{j = 1}^{s} \tr(C_j)}{\lambda_{\max}(K_O^{-1})} \leq \frac{1}{2} \sum_{j=1}^s |C_i|^{\tau/2} \leq \frac{1}{2} s^{1-\tau/2} \left(\sum_{j=1}^s |C_i| \right)^{\tau/2} \leq \frac{1}{2} s^{1-\tau/2} m^{\tau/2}.\]

We have the following additional requirement: \[s \leq \frac{m^{\frac{\tau}{2-\tau}}}{((1+\tau)\log m)^{2/(2-\tau)}}.\]
This condition controls the largest number of cliques; note that taking $\tau \rightarrow 1$ allows for nearly $m$ cliques (this is thus close to the case where all the sources are conditionally independent on the true label). Now, with a little bit of algebra, we have that
\begin{align*}
r_e(K_O^{-1}) &\leq \frac{1}{2} s^{1-\tau/2}m^{\tau/2} \leq \frac{m^{\tau/2}}{(1+\tau) \log m} m^{\tau/2} \nonumber \\
&= \frac{1}{2} \frac{m^{\tau}}{(1+\tau) \log m}.
\end{align*}

We will similarly require that $\|v\|$ is bounded by the expression above (that is, $O( \frac{1}{2} m^{\tau/2}/\log(m))$, so that 
\begin{align}
r_e(\Sigma_O) \leq  \frac{m^{\tau}}{(1+\tau) \log m}.
\label{eq:normal_cond}
\end{align}

Next, we have the alternative \emph{strong source block} condition. Now, we assume that one of the blocks corresponding to a clique is dominant. Concretely, if $C_i$ is dominant, we require that (i) $\tr(C_i) \geq \sum_{j \neq i} \tr(C_j)$, (ii) $\lambda_{\max}(C_i) \geq \lambda_{\max}(C_j)$ for all $j\neq i$, and (iii) $\|K_{OS}\|^2 \leq 2 \|(K_{OS})_{C_i}\|^2$. In the latter term, $(K_{OS})_{C_i}$ is the subvector of $K_{OS}$ corresponding to the variables in $C_i$. 

Under these assumptions, we show that the effective rank is bounded by a constant times $d$. First,
\[r_e(K_O^{-1}) = \frac{\tr(K_O^{-1})}{\|K_O^{-1}\|} = \frac{\tr(K_O^{-1})}{\lambda_{\max}(C_i)} \leq  \frac{2\tr(C_i)}{\lambda_{\max}(C_i)} \leq\frac{2d \lambda_{\max}(C_i)}{\lambda_{\max}(C_i)} = 2d.\]

Next, 
\begin{align*}
\|v\|^2/\|K_O^{-1}\| &=  \|cK_O^{-1}K_{OS}^T\|^2/\|K_O^{-1}\| \leq c^2 \|K_O^{-1}\| \|K_{OS}\|^2 \leq c^2 \|K_O^{-1}\| 2 \|(K_{OS})_{C_i}\|^2 \\
&\leq 2c^2 \|K_O^{-1}\| (d \|(K_{OS})_{C_i}\|_{\infty}^2) \leq c'd,
\end{align*}
for some constant $c'$. 

Putting these together, we have that 
\[r_e(\Sigma_O) \leq r_e(K_O^{-1}) + \|v\|^2/\|K_O^{-1}\| \leq 2d + c'd = c_4 d,\]
where $c_4 = 2+c'$, as desired.

\paragraph{Concentration Inequality} We use Proposition A.4 from \citet{Bunea15}. Written in our notation, it states that with probability at least $1-\exp(-t)$,
\begin{align}
\|\Sigma_O^{(n)} - \Sigma_O\| \leq c_2 \|\Sigma_O\| \max \left\{ \sqrt{ \frac{r_e(\Sigma_O) (t+\log m)}{n}}, \frac{r_e(\Sigma_O) ( t +  \log m)}{n} \right\}.
\label{eq:concbound1}
\end{align}
Note that this result applies to our setting, since our variables are indeed sub-Gaussian and have higher-order moments bounded in terms of the second moments. We can assume, without loss of generality, that our variables are centered (otherwise, we can estimate the mean from samples and produce a concentration bound at least as tight as the above). This enables us to use Proposition A.4.

Now, if $r_e(\Sigma_O) (t + \log m) / n \leq 1$, or, equivalently, $r_e(\Sigma_O) \leq \frac{n}{t +\log m}$, the max on the right hand side in \eqref{eq:concbound1} takes the first value. We can then rewrite the bound as
\begin{align}
\|\Sigma_O^{(n)} - \Sigma_O\| \leq c_2 \|\Sigma_O\| \sqrt{ \frac{r_e(\Sigma_O) (t+ \log m)}{n}}.
\label{eq:concbound2}
\end{align}

On the other hand, if  $r_e(\Sigma_O) > \frac{n}{t +\log m}$, we have that the second term in the max is larger, so that we obtain
\begin{align}
\|\Sigma_O^{(n)} - \Sigma_O\| \leq c_2 \|\Sigma_O\| \frac{r_e(\Sigma_O) (t+ \log(m))}{n} .
\label{eq:concbound3}
\end{align}

\paragraph{Remainder of the Proof}  Now we tackle the proof of the main theorem, which adapts the proof of Theorem 4.1 in \cite{Wu17}. The key is to replace Lemma D.1, which states (in our notation) that, for some constant $C_K$, 
\[\Pr \left\{ \|\Sigma_O^{(n)}-\Sigma_O\| \leq C_K \sqrt{ \frac{m}{n} } \right\} \geq 1 - 2 \exp(-m).\]

For the source block decay (SBD) assumption, we use \eqref{eq:concbound2} with $t = \tau \log m$. Then, as long as $n \geq m^{\tau}$, it is easy to verify that the condition \[r_e(\Sigma_O) \leq \frac{m^{\tau}}{\log m (m^{\tau})} \leq \frac{n}{(1+\tau)\log m}\] is met by directly applying the bound on $r_e(\Sigma_O)$ from \eqref{eq:normal_cond}. Next, 
\begin{align*}
\|\Sigma_O^{(n)} - \Sigma_O\| &\leq c_2 \|\Sigma_O\| \sqrt{ \frac{r_e(\Sigma_O) (1+ \tau) \log m}{n}} \\
&\leq c_2 \|\Sigma_O\| \sqrt{ \frac{m^{\tau}}{n}}.
\end{align*}
We write \[\mathcal{F}_{\text{sbd}}(n,m, \tau) =  c_2 \|\Sigma_O\| \sqrt{ \frac{m^{\tau}}{n}} = c_2 \psi_1 \sqrt{ \frac{m^{\tau}}{n}} .\]

Next, we work with the strong source block (SSB) condition. Again, we wish to obtain a final error probability of at least $1-m^{-\tau}$, so that we take $t = \tau \log m$. Applying our bound on $r_{e}(\Sigma_O)$,  the tail bound \eqref{eq:concbound3} becomes
\begin{align}
\|\Sigma_O^{(n)} - \Sigma_O\| \leq \frac{1}{n} c_2c_4 \|\Sigma_O\| d  (1 + \tau) \log(m).
\label{eq:concbound4}
\end{align}

We set
\[\mathcal{F}_{\text{ssb}}(n,m,d) :=  \frac{ c_2 c_4 \psi_1 d (1 + \tau) \log(m)}{n}.\]

Now that we have our two tail functions $\mathcal{F}_{\text{sbd}}(n,m, \tau)$ and $\mathcal{F}_{\text{ssb}}(n,m,d)$, we will finish off the proof by adapting the proof of \citet{Wu17}. For the first condition, we replace the tail term $\sqrt{m/n}$ with a $\sqrt{m^{\tau}/n}$ term, so that our require number of samples is the sublinear $m^{\tau}$ (rather than $m$).  For the second condition, we replace  $\sqrt{m/n}$ in \citet{Wu17} with a $(\log m) / n$ term, which produces a sampling rate in terms of $(\log m)$ instead of $m$.

Concretely, we replace the $C_K\sqrt{\frac{p}{n}}$ term in the proof of Theorem 4.1 in \citet{Wu17} with $\mathcal{F}_{\text{sbd}}(n,m,\tau) $ and  $\mathcal{F}_{\text{ssb}}(n,m,d)$. In particular, for the first case, we do this replacement in the following expression for the dual norm $g_{\gamma}$, 
\[g_{\gamma}(\mathcal{A}^\dagger h_{\Sigma^{(n)}_O}(R^*)) \leq m \|\Sigma^{(n)}_O R^*\| \leq \frac{ \gamma^{-1}}{\psi_m} \mathcal{F}_{\text{sbd}}(n,m,\tau),\]
in the step immediately preceding (D.4). Note that here, we replace $\max\{1, \gamma^{-1}\}$ with $\gamma^{-1}$, since in our regime of interest, $\gamma^{-1} \geq 1$. Indeed, this is equivalent to requiring that along with the condition on $\mu(\Omega)\xi(zz^T)$, we have $2d \geq \xi(zz^T)$. Note also that our notation for the smallest eigenvalue is slightly different.

The term then carries forward, with the final requirement being the selection of the regularization term $\lambda_n$ at the end of Step 1 of the proof. Hence, we now require that
\[\lambda_n = \frac{(3-2\nu) \gamma^{-1}}{\psi_m} \mathcal{F}(n,m,\tau)_{\text{sbd}}.\]

For the second condition, we replace the $C_K\sqrt{\frac{p}{n}}$ term with $\mathcal{F}_{\text{ssb}}(n,m,d) $. 
We now need that
\[\lambda_n = \frac{(3-2\nu) \gamma^{-1}}{\psi_m} \mathcal{F}(n,m,d)_{\text{ssb}}.\]

All that is left is to ensure the three conditions in the statement of Theorem 4.1 in \citet{Wu17} are met. These conditions are (in our notation)
\[\sigma > \frac{3}{\alpha} \lambda_n,\]
\[\frac{1}{\psi_m} > \frac{3\lambda_n}{\alpha},\]
and,
\[K_{O,\min} > \frac{3\gamma}{\alpha} \lambda_n.\]
Rewriting these, we have that
\begin{align}
\frac{1}{\lambda_n} > \frac{3}{\alpha} \max \left\{\frac{1}{\psi_m}, \frac{\gamma}{K_{O,\min}}, \sigma^{-1}\right\}.
\label{eq:threereq}
\end{align}

For our first condition, recalling that $\gamma = \frac{\nu \alpha}{2d\beta(2-\nu)}$,
\begin{align*}
\lambda_n &= \frac{(3-2\nu) \gamma^{-1}}{\psi_m} \mathcal{F}(n,m,\tau)_{\text{sbd}}\\
&=  \frac{2d \beta(3-2\nu)(2-\nu)}{\nu \alpha \psi_m} \mathcal{F}(n,m,\tau)_{\text{sbd}}\\
&=   \frac{2d c_2  \beta(3-2\nu \psi_1)(2-\nu)}{\nu \alpha \psi_m}  \sqrt{ \frac{m^{\tau}}{n}}.
\end{align*}
Then, plugging this into \eqref{eq:threereq}, we get
\[n > \left[ \frac{6 c_2  \beta(3-2\nu)(2-\nu)\psi_1}{\nu \alpha^2 \psi_m} \max \left\{\frac{1}{\psi_m}, \frac{\gamma}{K_{O,\min}}, \sigma^{-1}\right\} \right]^2 d^2 m^{\tau}.\]

This completes the first case of the theorem.

Now, for the second case,
\begin{align*}
\lambda_n &= \frac{(3-2\nu) \gamma^{-1}}{\psi_m} \mathcal{F}(n,m,\tau)_{\text{ssb}}\\
&=  \frac{2d \beta(3-2\nu)(2-\nu)}{\nu \alpha \psi_m} \mathcal{F}(n,m,\tau)_{\text{ssb}}\\
&=   \frac{2 c_2c_4 \beta(3-2\nu)(2-\nu)\psi_1 }{\nu \alpha \psi_m}  \frac{ d^2 (1 + \tau) \log(m)}{n}.
\end{align*}
Again, we plug the latter expression into \eqref{eq:threereq}, getting
\[n > \frac{2 c_2c_4 \beta(3-2\nu)(2-\nu)\psi_1 }{\nu \alpha^2 \psi_m} \max \left\{\frac{1}{\psi_m}, \frac{\gamma}{K_{O,\min}}, \sigma^{-1}\right\}  (1+\tau) d^2  \log(m).\]
Now we are done.
\end{proof}

\paragraph{Proof of Theorem~\ref{thm:lb_unsupervised}}
\begin{proof}
The typical approach taken for minimax-style lower bounds is to construct an ensemble of hypotheses (in our case, graphs encoding the distribution) and to control the distance between these hypotheses. Concretely, Fano's lemma is used, which requires controlling the KL divergence between pairs of distributions. As in prior work \cite{Santhanam12, Shanmugam14}, we use the symmetric KL divergence $S$, which is defined by \[S(f_G || f_{G'}) = D(f_G || f_{G'}) + D(f_{G'} || f_G),\]
with
\[D(f_G || f_{G'}) = \sum_{x \in \{0,1\}^m} f_G(x) \log \left( \frac{f_G(x)}{f_{G'}(x)} \right).\]

We use the following variant of Fano's lemma \cite{Santhanam12}
\begin{align}
n < (1-\delta) \frac{\log{M}}{\frac{2}{M^2} \sum_{k=1}^M \sum_{\ell=k+1}^M S(f_{G_k} || f_{G_\ell})} .
\label{eq:fano}
\end{align}

Here, we have a class of graphs $G_1, G_2, \ldots, G_M$. The result states that if $n$ is upper bounded as in \eqref{eq:fano}, \emph{no} structure learning procedure has a better maximum error probability (over the entire family) than $\delta - \frac{1}{\log M}$. Prior work on lower bounding the sample complexity for structure learning uses multiple choices of ensemble and takes the maximum over the resulting complexities. In particular, \citet{Santhanam12} (called SW from now on), considers three ensembles. The first of these takes a graph on $m$ nodes with no edges, and then adds one edge to form $\binom{M}{2}$ graphs. We will use a similar construction, with the additional constraint that we are in the weak supervision setting, where we have the label node $\y$ connected to all other nodes. 

We start with full generality. Note that, from our model, 
\begin{align*}
f_G(\lambda_1, \ldots, \lambda_m) &= \sum_{y} f_G(\lambda_1, \ldots, \lambda_m, y) \\
&= \sum_{y} \frac{1}{Z}\exp \left( \sum_{\lambda_i \in V} \theta_{i} \lambda_i + \sum_{(\lambda_i,\lambda_j)\in E} \theta_{ij} \lambda_i \lambda_j  + \theta_Y y + \sum_{\lambda_i \in V} \theta_{Y,i} y \lambda_i \right) \\ 
&=  \frac{1}{Z}\exp \left( \sum_{\lambda_i \in V} \theta_{i} \lambda_i + \sum_{(\lambda_i,\lambda_j)\in E} \theta_{ij} \lambda_i \lambda_j \right) \left[ \sum_y \exp \left(  \theta_Y y + \sum_{\lambda_i \in V} \theta_{Y,i} y \lambda_i \right) \right].
\end{align*}

Now we can start computing the symmetric KL divergence between a pair of graphs $G, G'$ in our class of graphs:
\begin{align}
S(G || G') &=  \mathbb{E}_{G}[\log f_G - \log f_{G'}] + \mathbb{E}_{G'}[\log f_{G'} - \log f_{G}] \\ \nonumber
&= \mathbb{E}_{G}\left[ \sum_{\lambda_i \in V} (\theta_i - \theta'_i) \lambda_i + \sum_{(\lambda_i,\lambda_j)\in E} (\theta_{ij} - \theta'_{ij}) \lambda_i \lambda_j + \log \frac{\sum_y \exp( \theta_Y y + \sum_{\lambda_i \in V} \theta_{Y,i} y \lambda_i  )}{\sum_y \exp(\theta'_Y y + \sum_{\lambda_i \in V} \theta'_{Y,i} y \lambda_i ) }\right] \\ \nonumber
&\quad+  \mathbb{E}_{G'}\left[ \sum_{\lambda_i \in V} (\theta'_i - \theta_i) \lambda_i + \sum_{(\lambda_i,\lambda_j)\in E'} (\theta'_{ij} - \theta_{ij}) \lambda_i \lambda_j + \log \frac{\sum_y \exp( \theta'_Y y + \sum_{\lambda_i \in V} \theta'_{Y,i} y \lambda_i  )}{\sum_y \exp(\theta_Y y + \sum_{\lambda_i \in V} \theta_{Y,i} y \lambda_i ) }\right] \\
\label{eq:kl_general}
\end{align}
Here, the partition functions cancel out going from the first line to the second.

Now we build our ensemble. Let $G^{st} = (V,E)$, with $V= \{\lf_1, \ldots, \lf_m, \y\}$. We set \[E = \{ (\lf_s \lf_t), (\lf_1, \y), (\lf_2, \y), \ldots, (\lf_m, \y)\}.\]
Note that the edges consist of the latent label node connected to all other nodes, and the sole additional edge $(\lf_s, \lf_t)$. 

For this class of models, we consider only edge potentials, all with parameter $\theta$ (of course, the non-edges have a parameter of 0). With this setting,  for two graphs $G^{st}, G^{uv}$, where the edge sets are $E$ and $E'$, respectively, \eqref{eq:kl_general} reduces to, 
\begin{align*}
S(G^{st} || G^{uv}) &= \mathbb{E}_{G^{st}}\left[  \sum_{(\lambda_i,\lambda_j)\in E, \not\in E'} \theta \lambda_i \lambda_j -  \sum_{(\lambda_i,\lambda_j)\in E', \not\in E} \theta \lambda_i \lambda_j + \log \frac{\sum_y \exp(  \sum_{\lambda_i \in V} \theta y \lambda_i  )}{\sum_y \exp( \sum_{\lambda_i \in V} \theta y \lambda_i ) }\right] \\
&\quad+  \mathbb{E}_{G^{uv}}\left[ \sum_{(\lambda_i,\lambda_j)\in E' \not\in E} \theta \lambda_i \lambda_j -  \sum_{(\lambda_i,\lambda_j)\in E, \not\in E'} \theta \lambda_i \lambda_j+ \log \frac{\sum_y \exp(  \sum_{\lambda_i \in V} \theta y \lambda_i  )}{\sum_y \exp( \sum_{\lambda_i \in V} \theta y \lambda_i ) }\right] \\
\end{align*}
Note that the fraction inside the log is equal to 1---this is because there is an edge between $\lf_i$ and $\y$ for all $i$. As a result, the log term is 0. Note also that there is only one edge that differs in each graph, so that the above reduces further to 
\[ S(G^{st} || G^{uv}) =  \theta (\mathbb{E}_{G^{st}}[\lf_s \lf_t]-\mathbb{E}_{G^{uv}}[\lf_s \lf_t]) + \theta  (\mathbb{E}_{G^{uv}}[\lf_u \lf_v]-\mathbb{E}_{G^{st}}[\lf_u \lf_v])  .\]
By symmetry, this is simply
\begin{align}
 S(G^{st} || G^{uv}) =  2\theta  (\mathbb{E}_{G^{st}}[\lf_s \lf_t]-\mathbb{E}_{G^{uv}}[\lf_s \lf_t])   
\label{eq:symm}
\end{align}

In the supervised case, in the above expression there is no path connecting $\lf_s$ to $\lf_t$ in $G^{uv}$, so that $\mathbb{E}_{G^{uv}}[\lf_s \lf_t] = 0$ and the result further reduces to $2\theta  (\mathbb{E}_{G^{st}}[\lf_s \lf_t]$. In particular, a simple computation shows that this is equal to $2\theta \tanh \theta$. Hower, in the supervised case, $\mathbb{E}_{G^{uv}}[\lf_s \lf_t]) \neq 0$, since \emph{there is a path} between $\lf_s$ and $\lf_t$ despite the fact that in $G^{uv}$ there is no edge joining them! The path is through the latent variable $\y$: $\lf_s - \y  - \lf_t$. As a result, the $\mathbb{E}_{G^{uv}}[\lf_s \lf_t]) > 0$ and this term \emph{reduces} the overall distance between our graphs. In turn, this means that we require \emph{more} samples for the weak supervision case compared to the supervised setting. We make these notions concrete in the following. 

We compute $\mathbb{E}_{G^{st}}[\lf_s \lf_t]$ and $\mathbb{E}_{G^{uv}}[\lf_s \lf_t]$. This is a simple calculation. Note that since we marginalize over the latent $\y$, the vertices $\lf_w$ for $w \not\in \{\lf_s, \lf_t\}$ do not contribute anything. Then, we have that
\[\mathbb{E}_{G^{st}}[\lf_s \lf_t] = \frac{\exp(3 \theta) - \exp(-\theta) }{\exp(3\theta) + 3 \exp(-\theta)},\]
and
\[\mathbb{E}_{G^{uv}}[\lf_s \lf_t] = \frac{\exp(2 \theta) +\exp(-2\theta) - 2 }{\exp(2\theta) + \exp(-2\theta) + 2} = \tanh^2(\theta),\]
A small simplification to the first term and plugging this into \eqref{eq:symm} yields
\[S(G^{st} || G^{uv}) = 2\theta(1 - 4(\exp(4\theta)+3)^{-1} - \tanh^2(\theta)).\]

Finally, we are ready to apply Fano \eqref{eq:fano}. Since we have $\binom{m}{2}$ choices for which edge to choose, and since $S(G^{st}||G^{uv})$ is the same for all choices, we obtain the bound 
\[ n  < (1-\delta) \frac{\log (m(m-1)/2)}{2 \theta (1 - 4(\exp(4\theta)+3)^{-1} - \tanh^2(\theta))}.\]
This completes the proof.
\end{proof}

\paragraph{Conjecture on Singleton Separator Set Dense Ensemble}

Our proof above used a \emph{sparse ensemble} to derive a lower bound: such ensembles use few edges between the labeling functions, and add an edge. Note that our construction satisfied the singleton separator set property. The other approach is to consider \emph{dense ensembles}, as in the second ensemble in SW.

This particular ensemble involves dividing up the $m$ nodes evenly into cliques of $d$ nodes each. Then, a single edge is removed from one of these cliques. 
However, this ensemble \emph{does not} satisfy the singleton separator set assumption. To see why, suppose $\lf_1, \ldots, \lf_d$ is such a clique, and remove the edge between $\lf_i,\lf_j$ for $1 \leq i,j \leq d$. Now, we have two maximal cliques: $\{\lf_1, \ldots, \lf_d\} \setminus \{\lf_i\}$ and $\{\lf_1, \ldots, \lf_d\} \setminus \{\lf_j\}$. The separator set is the intersection of these two cliques: $\{\lf_1, \ldots, \lf_d\} \setminus \{\lf_i, \lf_j\}$, which is not a singleton for $d>3$.  

For the second ensemble, the idea is to ensure that $\mathbb{E}[\lf_i, \lf_j]$ is very close to 1, e.g., at least $1-\exp(d\theta)/d$. We conjecture that the following ensemble, which \emph{does} satisfy the singleton separator set property, also has the same behavior. Specifically, take two complete graphs $K_{d+1} \cup K_{d+1}$. Let us name the vertices as $\{\lf_i\} \cup \{\lf_{d+i}\}$ for $1 \leq i \leq d$. Now, in addition to the two complete cliques, we also add cross-edges (acting as ribs), $(\lf_i, \lf_{d+i})$ for $1 \leq i \leq d$. Now we remove a single edge, say $(\lf_s, \lf_{d+s})$ to form $G^s$. Note that $G^s$ does satisfy singleton separator set: the maximal cliques are the two complete subgraphs, along with each of the cross edges. The intersections here are the single nodes that connect the clique with the cross-edge.

We conjecture that using $G^s$ as in Ensemble 2 in SW will give us a similar amount of control over  $\mathbb{E}[\lf_s, \lf_{d+s}]$, providing us with a similar bound for the more restricted singleton separator set ensemble. Deriving such a bound would present another bounding regime, sharpening our result in Theorem~\ref{thm:lb_unsupervised}.

\paragraph{Beyond Singleton Separator Sets}
There is a further remarkable application of robust PCA. Say we wish to perform structure learning (in the fully supervised setting, where we see $\Sigma$) by using a covariance matrix-based approach, but our graph $G$ \emph{does not} satisfy the singleton separator set assumption. Then, $\Sigma^{-1}$ is not graph-structured, but an enlarged \emph{generalized covariance matrix} $\Sigma_{\text{aug}}$ is, where this matrix is augmented with variables in the separator set $S$ \cite{Loh13}. Then robust PCA can recover the structure with only $\Sigma^{-1}$ as input.

More concretely, suppose that $G$ is a graph where all the separator sets are singleton, with the exception of one set $\{\lf_s, \lf_t\}$. Then, the generalized covariance matrix contains all variables and an additional row corresponding to the entry $\lf_s \lf_t$. Note that here we observe all the $\lf_i$'s, but, as we do not know the structure, we cannot select $\lf_s \lf_t$ to form the generalized covariance matrix. However, if we treat this variable as \emph{latent}, taking the role of $\y$ in our analysis, we can use Algorithm~\ref{alg:sl1}. In particular, if our second condition is met, our sample complexity is again $\Omega(d^2 \log m)$, which extends the result in \citet{Loh13} to the non-singleton separator set graph class.

%% file: ext.tex
For Algorithm~\ref{alg:sl1}, cases where we have direct access to the inverse matrix $\Sigma_O^{-1}$, such as in \cite{Ratner19}, where it is computed for the parameter estimation algorithm, the loss function term in Algorithm~\ref{alg:sl1} is not needed, and we simply run robust PCA.
We now provide additional comparisons to a structure inference method for weak supervision, \citet{varma2017inferring}. The method uses the source code that defines different weak supervision sources to infer dependencies among them by looking at what features of the data the sources rely on. This approach has an advantage over statistical methods since it does not require \emph{any} data to infer partial structure. However, this method is unable to infer any structure for CDR and IMDb, performing up to $3.91$ F1 points worse than our method.
For the other tasks, our method is able to learn dependencies that \citet{varma2017inferring} infers \emph{and} additional dependencies that are implicit for the Bone Tumor and MS-COCO tasks. This leads to an improvement of up to $1.60$ F1 points as shown in Table~\ref{table:results_ext}.

\begin{table*}[t]
    \centering    
    \begin{tabular}{@{}lccccccccc@{}}
        \toprule
                             & \multicolumn{1}{l}{} & \multicolumn{1}{l}{} & \multicolumn{1}{l}{} & \multicolumn{1}{l}{} & \multicolumn{1}{l}{} & \multicolumn{1}{l}{} & \multicolumn{1}{l}{} & \multicolumn{2}{c}{\textbf{Improvement Over}} \\ \cmidrule(l){9-10} 
        \textbf{Application} & \textbf{$m$}      & \textbf{($s,d$)}  & \textbf{MV}          & \textbf{Indep.}      & \textbf{Bach et al.} & \textbf{Partial}     & \textbf{Ours}        & \textbf{Bach et al.}    & \textbf{Partial}    \\ \midrule
        Bone Tumor           & 17                    & (2,3)                    & 65.72                & 67.32                & 67.83                & 70.79                & 71.96                & +4.13                   & +1.17               \\
        CDR                  & 33                   & (22,14)                  & 47.74                & 54.60                & 55.90                & 54.60                & 56.81                & +0.91                    & +2.21               \\
        IMDb                 & 5                    & (1,4)                    & 55.21                & 58.80                & 60.23                & 58.80                & 62.71                & +2.48                   & +3.91               \\
        MS-COCO              & 3                    & (1,2)                    & 57.95                & 59.47                & 59.47                & 62.28                & 63.88                & +4.41                   & +1.60              \\ \bottomrule
        \end{tabular}
    \caption{Extended results table with comparison to the structure inference method (Partial).}
    \label{table:results_ext}
    \end{table*}